\documentclass[11pt,a4paper]{article}

\usepackage[T1]{fontenc}
\usepackage[utf8]{inputenc}
\usepackage{lmodern}
\usepackage[margin=2.5cm]{geometry}
\usepackage{graphicx}
\graphicspath{{figures/}{../figures/}}
\usepackage{booktabs}
\usepackage[round,authoryear]{natbib}
\usepackage{xcolor}
\usepackage{xurl}
\usepackage[colorlinks=true,linkcolor=blue!50!black,citecolor=blue!50!black,urlcolor=blue!50!black]{hyperref}

\newcommand{\dfine}{D-FINE-seg}

\title{Revisiting RadioGalaxyNET: modern detectors for radio galaxy detection and host identification}
\author{Argo Saakyan\\{Agnify.ai, argo.cve@gmail.com}}
\date{}

\begin{document}
\maketitle

\begin{abstract}
RadioGalaxyNET is the public benchmark for finding extended radio galaxies in EMU pilot-survey images, classifying them and locating their infrared hosts. Its best published results date from 2023: 54.5 box AP (Gal-YOLOv8) and 48.1 host keypoint AP (Gal-DINO). We train three current detectors, D-FINE-seg, YOLO26-pose and RF-DETR, on the same data and score them with the benchmark's own evaluation code, which reproduces the published Gal-DINO numbers exactly. All exceed the published results. D-FINE-seg M reaches 67.1 box AP and 72.4 host AP, 12.6 and 24.3 points above the best published results, and the highest host AP of the detectors we tested; on boxes it is level with YOLO26. Its host lead over YOLO26 comes from host-centered images, which make up most of the benchmark. End to end, D-FINE-seg S is the fastest model at 1.9 ms per image; engine against engine, YOLO26 is faster at the same size. We also report what matters for anyone using the benchmark: its scoring protocol adds 5 to 10 AP, its host labels sit at the image center, its box labels are offset from the image by 1.5 pixels, and its images can be rebuilt from the public survey data, which recovers sky positions and doubles as extra training data. We suggest that future work report host identification F1, as a measure closer to catalog use: D-FINE-seg M 71.8, YOLO26-pose M 70.5, RF-DETR 63.8, Gal-DINO 59.2.
\end{abstract}

\section{Introduction}

The Evolutionary Map of the Universe \citep[EMU;][]{norris2021}, a survey with the Australian Square Kilometre Array Pathfinder \citep[ASKAP;][]{hotan2021}, will detect millions of extended radio galaxies, and each needs a morphological class and an infrared host galaxy. RadioGalaxyNET \citep{gupta2024rgnet} made this a computer-vision benchmark: 2{,}799 images that combine an EMU pilot-survey radio cutout with an infrared cutout of the same field from the Wide-field Infrared Survey Explorer \citep[WISE;][]{wright2010} at 3.4~$\mu$m, labeled with a box and class (FR-I, FR-II, FR-X or R) for each extended source and a keypoint for its host. The original paper trained several detectors; the best are Gal-YOLOv8 for boxes (54.5 AP) and Gal-DINO, a DINO detection transformer \citep{zhang2023dino} with a keypoint branch, for hosts (48.1 AP). The only later result we found, LARGAL \citep{anand2026largal}, reports 53.5 box AP with its own evaluation script.

Real-time detectors have improved considerably since 2023. We ask how far current off-the-shelf models move this benchmark, using the benchmark's own scoring, and what a user of the benchmark should know when comparing against it.

\section{Setup}

\begin{figure}[t]
\centering
\includegraphics[width=\textwidth]{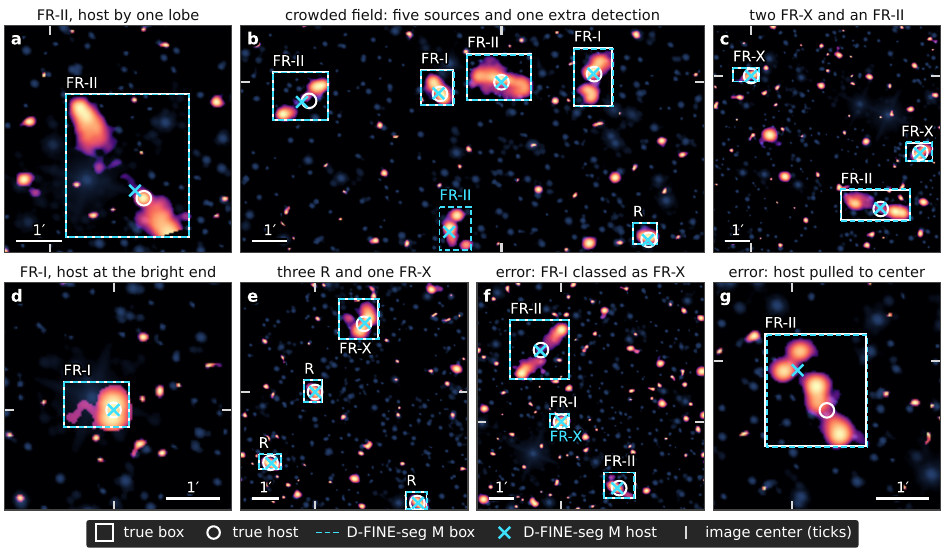}
\caption{Crops of test images with their labels (white: box, class and host) and the predictions of \dfine{} M (cyan; seed 42, class-agnostic NMS at IoU 0.7 and confidence $\geq 0.72$, as in Table~\ref{tab:hostid}). Radio emission is shown in orange and purple over WISE 3.4\,$\mu$m infrared in blue; ticks on the frame mark the image center. (a--e) Every labeled source is identified, including hosts far from the center of their box (a, d) and away from the image center (a, b, c, e); the extra detection in (b), marked with its predicted class, has no label and may be an unlabeled source. (f) A correct box and host with the wrong class. (g) The only test image without a host at its center: \dfine{} finds the box but places the host near the image center.}
\label{fig:examples}
\end{figure}

\paragraph{Data and scoring.} We use the public release \citep{rgnetdata}: 1{,}959 training, 420 validation and 420 test images of $450\times450$ pixels, with the original splits. All results are on the test split (611 sources). Figure~\ref{fig:examples} shows examples. We score every model with Gal-DINO's evaluation: COCO box AP and keypoint AP \citep{lin2014coco} with the benchmark's size ranges and OKS $\sigma = 0.107$. Our stand-alone re-implementation gives 53.73 box AP and 48.14 keypoint AP for Gal-DINO's released checkpoint, identical to the paper and to Gal-DINO's own evaluation script. As in Gal-DINO's evaluation and the usual COCO protocol for detection transformers, each model submits its 300 highest-scoring (box, class) pairs per image without non-maximum suppression (NMS); Section~\ref{sec:notes} shows the effect of this choice. Every model runs through its own framework's inference and post-processing code.

\paragraph{Models.} \dfine{} \citep{saakyan2026dfineseg}, developed by the author, is an open-source framework built on the D-FINE detector \citep{peng2024dfine} that adds segmentation and keypoint heads. Its keypoint head predicts each keypoint as an offset from the box center after every decoder layer, refines it with deformable attention to the image features and predicts the keypoint's quality (its expected OKS); the top detections are ranked by the geometric mean of class score and this quality. YOLO26-pose is the current Ultralytics pose model \citep[version 8.4.92;][]{ultralytics}. RF-DETR \citep{rfdetr} is a real-time detection transformer whose only released keypoint model is an XL preview at 576 pixels. All predict one keypoint per object, the host. We train \dfine{} S, M, L and X, YOLO26-pose S, M, L and X, and RF-DETR XL.

\paragraph{Training.} All models start from COCO weights and train and run at their framework's default input size (640 pixels for \dfine{} and YOLO26, 576 for RF-DETR; the 450-pixel images are upsampled). \dfine{} uses its default recipe for 75 epochs with horizontal and vertical flips and $90^\circ$ rotations, the benchmark's OKS $\sigma$ in its keypoint loss, and no photometric augmentation: the images store the radio as 16-bit values split over two 8-bit channels (blue holds the value divided by 256, green the remainder), so a brightness or contrast change would scramble it. For YOLO26 we tried several recipes and report the best one on the validation split: the Ultralytics defaults for 200 epochs, which flip images horizontally but not vertically. Vertical flips cost YOLO26 25 box AP because of an offset in the box labels (Section~\ref{sec:notes}); with the offset removed from the training boxes and added back to the predictions, YOLO26-pose M trained with vertical flips scores 65.6 box and 61.9 host AP, no better than without them (67.7 and 62.3 over three seeds), so we keep the defaults. Training YOLO26 with the benchmark's OKS $\sigma$ in its keypoint loss, instead of the Ultralytics default, lowered host AP (57.2 against 61.0 for S). RF-DETR uses its default keypoint recipe for 100 epochs with one change: the weight of its keypoint uncertainty loss is set to 0 (default 1.0). With the default, that loss has no lower bound: each cutout was made by centering it on the host galaxy of one source, so in 2{,}789 of the 2{,}799 images one host keypoint sits exactly at the image center. The model can predict that point almost exactly and keep lowering the loss by growing more confident, which crowds out the classification loss, and the classifier never trains (15.4 box AP). We also keep RF-DETR's horizontal flip, which it otherwise drops for single-keypoint data. Each model keeps its best epoch on validation.

\paragraph{Latency.} All models run as TensorRT fp16 engines at batch 1 on an idle RTX 5070 Ti, back to back in one session. End-to-end time is one call of each framework's own inference code, from a decoded image in memory to final boxes and keypoints on the CPU; \dfine{} and Gal-DINO replay CUDA graphs, and \dfine{}'s time includes its NMS (about 0.1\,ms). Engine time is the TensorRT engine alone, launched the same way for every model, without a CUDA graph, which is why Gal-DINO's engine time slightly exceeds its end-to-end time. Replaying each engine as a CUDA graph saves 0.2 to 0.5\,ms for every model and does not change their order. RF-DETR's engine is built by its own export tool with TensorRT 10.11, the others with 10.13.

\section{Results}

\begin{table}[t]
\centering
\caption{RadioGalaxyNET test split: box AP and host keypoint AP (\%, COCO AP@[.5:.95]), Gal-DINO's evaluation and protocol for all models, PyTorch weights. Rows with $\pm$ are means and sample standard deviations over three training seeds; the other trained rows are single runs. Resampling the test images moves any single model's AP by about 2 points (one standard deviation). Latency in ms, TensorRT fp16, batch 1, RTX 5070 Ti: end to end (e2e) through each framework's own code, and engine only. Parameters are those of the deployed (fused) model.}
\label{tab:main}
\small
\begin{tabular}{lrrrrrrr}
\toprule
& & \multicolumn{2}{c}{box} & \multicolumn{2}{c}{host keypoint} & \multicolumn{2}{c}{ms} \\
\cmidrule(lr){3-4}\cmidrule(lr){5-6}\cmidrule(lr){7-8}
model & params & AP & AP$_{50}$ & AP & AP$_{50}$ & e2e & engine \\
\midrule
Gal-YOLOv8 (published) & & 54.5 & 58.5 & -- & -- & & \\
Gal-DINO (published) & 47\,M & 53.7 & 60.2 & 48.1 & 53.4 & 11.6 & 11.7 \\
\midrule
YOLO26-pose S & 9.8\,M & 63.0$\pm$2.8 & 68.5 & 61.0$\pm$0.3 & 66.1 & 2.4 & \textbf{1.1} \\
YOLO26-pose M & 21.2\,M & \textbf{67.7}$\pm$0.9 & 71.0 & 62.3$\pm$1.3 & 68.0 & 3.0 & 1.6 \\
YOLO26-pose L & 25.6\,M & 65.7 & 68.9 & 60.3 & 66.0 & 3.5 & 2.2 \\
YOLO26-pose X & 57.6\,M & 63.9 & 67.7 & 58.0 & 64.3 & 4.9 & 3.6 \\
\midrule
RF-DETR XL & 40.7\,M & 57.3 & 62.8 & 56.6 & 60.5 & 5.4 & 2.4 \\
\midrule
\dfine{} S & 13.4\,M & 66.4$\pm$0.7 & 73.7 & 70.9$\pm$0.4 & 72.6 & \textbf{1.9} & 1.6 \\
\dfine{} M & 22.8\,M & 67.1$\pm$1.0 & \textbf{75.5} & \textbf{72.4}$\pm$1.2 & \textbf{74.3} & 2.4 & 2.2 \\
\dfine{} L & 34.8\,M & 65.2 & 73.1 & 68.0 & 71.4 & & \\
\dfine{} X & 66.2\,M & 62.6 & 70.5 & 65.8 & 68.9 & 4.1 & 4.1 \\
\bottomrule
\end{tabular}
\end{table}

Table~\ref{tab:main} gives the results. Every current model beats the published results, \dfine{} and YOLO26 by a wide margin: \dfine{} M by 12.6 box AP over Gal-YOLOv8 and 24.3 host AP over Gal-DINO. The test split is small: resampling its 420 images (paired bootstrap) moves any model's AP by about 2 points, so two models must differ by about 4 AP to be told apart at 95\% confidence.

On boxes, \dfine{} M and the best YOLO26, M, are level: \dfine{} minus YOLO26 is $-0.6$ AP, with a 95\% interval of $-4.8$ to $+2.9$; \dfine{} finds more of the sources (AP$_{50}$ 75.5 against 71.0) and YOLO26 places its boxes more tightly. On hosts, \dfine{} leads YOLO26 by 8 to 10 AP at equal size (M: 95\% interval $+6.6$ to $+13.4$). With about 2{,}000 training images, larger models do not help: \dfine{} L and X score below S and M, and YOLO26-pose L and X below M. RF-DETR XL, the only released RF-DETR keypoint model, is below every YOLO26 and \dfine{} size. Trained as box detectors alone (three seeds each; \dfine{} with the best epoch chosen on box AP, YOLO26 with its default recipe), \dfine{} M reaches 71.2 box AP and YOLO26 M 69.4, and the two remain level: \dfine{} leads by 1.9 on test (95\% interval $-1.1$ to $+5.1$), YOLO26 by 3.1 on validation.

End to end, \dfine{} S is the fastest model, and \dfine{} M matches YOLO26-pose S at 2.4\,ms with 4 more box and 11 more host AP. Engine against engine, YOLO26 is faster at the same size; \dfine{}'s end-to-end lead comes from its leaner inference code. TensorRT engines score within 0.35 AP of PyTorch for every model.

\paragraph{Where the host lead comes from.} 419 of the 611 test hosts sit at exactly the image center (Section~\ref{sec:notes}). On these, \dfine{} M leads YOLO26-pose M by 13 host AP (76.5 against 63.5, means of three seeds). On the 192 off-center sources the two are level on hosts (32.7 and 30.7) and YOLO26 is better on boxes (44.4 against 33.2). Over all sources, \dfine{}'s lead grows from 6.2 host AP at the loosest keypoint match (OKS 0.5) to 17.4 at the strictest (OKS 0.95): it finds somewhat more hosts and places them more tightly.

\paragraph{Host identification.} AP summarizes a ranking. Table~\ref{tab:hostid} instead uses the models as a catalog would: class-agnostic NMS so that each source gets one class, one operating point per model, and a count of the sources identified. \dfine{} M has the highest precision and F1, 71.8, but YOLO26-pose M is close (\dfine{} minus YOLO26: $+1.3$, 95\% interval $-1.1$ to $+3.8$) and has the highest recall; both are well ahead of RF-DETR and Gal-DINO (\dfine{} minus Gal-DINO: $+12.7$, interval $+9.2$ to $+16.0$). Off-center, YOLO26 identifies the most sources, as it is also best on off-center boxes. Without NMS, every F1 changes by less than one point and the order is the same.

\begin{table}[t]
\centering
\caption{Host identification on the test split, at each model's confidence threshold chosen for box F1 on validation. A source is identified when a detection has its class, box IoU $\geq 0.5$ and a host within OKS 0.5 of the true one; every other kept detection is a false identification. Precision: identified sources per kept detection. Recall: identified share of the 608 test sources whose host lies inside the image (off-center: of the 189 whose host is away from the image center). \dfine{} and YOLO26: means of three seeds. Class-agnostic NMS at IoU 0.7 on validation and test, before the threshold is chosen. PyTorch weights, except Gal-DINO (its TensorRT fp16 engine).}
\label{tab:hostid}
\small
\begin{tabular}{lrrrr}
\toprule
model & precision & recall & F1 & off-center recall \\
\midrule
\dfine{} M & \textbf{71.4} & 72.3 & \textbf{71.8} & 61.6 \\
YOLO26-pose M & 67.4 & \textbf{74.0} & 70.5 & \textbf{73.2} \\
RF-DETR XL & 61.7 & 66.0 & 63.8 & 53.4 \\
Gal-DINO & 61.0 & 57.4 & 59.2 & 28.6 \\
\bottomrule
\end{tabular}
\end{table}

\section{Notes for users of the benchmark}
\label{sec:notes}

\paragraph{Multi-label submission adds 5 to 10 AP.} Without NMS, one box can be submitted under several classes, and COCO AP scores each class separately, which raises recall wherever the class is ambiguous, as it often is between FR-I, FR-II and FR-X. Keeping only the best class of each box gives nearly the same numbers as class-agnostic NMS at IoU 0.7 (\dfine{} M, three seeds: 58.7 against 58.5 box AP), so the gain comes from the extra classes, not from duplicate boxes. NMS 0.7 lowers every model by 5 to 10 AP (\dfine{} M 58.5 box / 62.7 host AP, YOLO26-pose M 61.4 / 56.6, Gal-DINO 44.1 / 39.5; PyTorch, three seeds for the M models). The host order holds; on boxes YOLO26 M and L move ahead, M by 2.9 AP. Gal-YOLOv8's published number used Ultralytics' validation, which also allows several classes per box; \dfine{} S scores 67.0 under that exact protocol, so its comparison with 54.5 is like for like. New results should state which protocol they use.

\paragraph{Hosts sit at the image center.} The host keypoint is at the image center in 2{,}789 of the 2{,}799 images, and models learn this: replacing every predicted host of \dfine{} M with the image center still gives 51 host AP, above the published 48.1. Figure~\ref{fig:examples}g shows the pull of the center on the one test image without a centered host. On 109 test fields rebuilt from raw data (below) and centered on the CatWISE host position instead, 0--6 pixels away, \dfine{} S's host AP falls from 82.9 to 76.5. Keypoint AP on this benchmark therefore overstates performance on cutouts that are not host-centered, such as a sliding window over a survey; reporting off-center sources separately as well shows how much of a result depends on the centered labels.

\paragraph{Box labels are offset from the image.} Each box label sits 1.5 pixels above and to the left of the radio emission it encloses (median over the boxes clear of the image edge, the same in the training and test splits); the host keypoints match the infrared host to within 0.1 pixel. The error is in the labels: 97\% of the released polygon vertices are vertices of scikit-image contours of the radio emission moved one pixel up and to the left, stored without the half-pixel shift to COCO's pixel-corner convention, which adds the other 0.5 pixel; each box is its polygon's extent. A flip reverses this offset relative to the image content, so flip augmentation creates contradictory targets unless the model can tell how the image was flipped. A YOLO26s detector trained with horizontal flips only scores 64.2 box AP on the test images but 20.1 on the same images flipped vertically; trained with vertical flips as well, it places its boxes on the emission, 1.5 pixels from the labels, and scores 39. With the boxes moved onto the emission for training, YOLO26-pose M with vertical flips scores 65.6 box AP once the offset is added back to its predictions and 43.3 without it, at the same AP$_{50}$: a 1.5-pixel shift alone costs 22 box AP, because the stricter IoU thresholds of COCO AP fail on these small boxes. \dfine{}, trained with vertical flips and $90^\circ$ rotations, places its boxes within 0.5 pixels of the labels. New work should correct the boxes before using flips or rotations, and add the offset back before scoring.

\paragraph{Small data quirks.} The release has 2{,}799 images (the paper quotes 2{,}800). Seventeen host keypoints lie outside their image, three of them in the test split. One training image is identical to a validation image. The splits are by image, not by sky region: of the 361 test fields we could place on the sky (below), 275 overlap a training field, and in 118 the test host lies inside a training image. None of this changes the comparison between models, but all of it changes exact numbers.

\paragraph{The images can be rebuilt from the survey data.} The release has no sky coordinates. The first six digits of each file name match the declination of the image center (DDMMSS, inferred from the data), which with CatWISE2020 \citep{marocco2021} and the radio emission pattern locates each field. From the native-resolution EMU-PS1 mosaic and unWISE W1 \citep{lang2014}, with the published preprocessing \citep{gupta2023galcam}, we rebuild 2{,}388 of the 2{,}799 images, with a median mask overlap against the released ones of 0.83 for the radio and 0.85 for the infrared. On the 355 rebuilt test images, host AP holds and box AP drops (\dfine{} S 69.7 host / 59.8 box AP against 71.2 / 66.8 on the same released images; Gal-DINO 50.3 / 52.5 against 49.8 / 55.7), and the two models keep their order. Adding the rebuilt training images as a second rendering of the same galaxies raises \dfine{} S on the original test split from 66.4 to 70.4 box AP and from 70.9 to 74.0 host AP (three seeds). This uses data the published models did not see, so we keep it out of Table~\ref{tab:main}.

\section{Conclusion}

D-FINE-seg and YOLO26 beat the published RadioGalaxyNET results by a wide margin. \dfine{} M achieves the highest host AP of the models we tested and is level with YOLO26 on boxes. Most of \dfine{}'s host-AP lead over YOLO26 comes from the centered host labels; on off-center hosts the two are level, and that subset, scored against corrected box labels, is a useful second view of future results. Host-identification F1 is closer to catalog use, since a catalog runs a model at a fixed threshold with one class per source. On it, \dfine{} is slightly ahead of YOLO26, within noise, and both are well ahead of Gal-DINO.

\section*{Data and code availability}

RadioGalaxyNET is available from the CSIRO Data Access Portal \citep{rgnetdata}; EMU-PS1 from CASDA; unWISE and CatWISE2020 from IRSA. The code to reproduce every result is at \url{https://github.com/ArgoHA/radiogalaxynet-dfine-vs-galdino}: the evaluation, training and inference code and configurations for all three trained frameworks, the analyses, the code that rebuilds the images from the survey data, and the figure scripts. The repository also bundles the predictions of \dfine{} S and Gal-DINO. The \dfine{} S and M weights (the seed-42 runs of Table~\ref{tab:main}; test box / host AP: S 67.2 / 70.4, M 68.0 / 71.1) are at \url{https://huggingface.co/ArgoSA/dfine-seg-radio-galaxy-net}.

\section*{Competing interests}

The author is the first author and a developer of \dfine{} \citep{saakyan2026dfineseg}, one of the frameworks compared here. To keep the comparison fair, every model is scored with the benchmark's own evaluation code, each framework is trained and run with its own code, and the code and configurations behind every result are public.

\section*{Use of AI tools}
Code for the model comparison and a first draft were produced with AI assistance under the author's direction; the author verified all results and takes full responsibility.

\section*{Acknowledgements}

We thank the RadioGalaxyNET and Gal-DINO authors for releasing their data, code and checkpoint. This scientific work uses data obtained from Inyarrimanha Ilgari Bundara, the CSIRO Murchison Radio-astronomy Observatory. We acknowledge the Wajarri Yamaji People as the Traditional Owners and native title holders of the Observatory site. CSIRO's ASKAP radio telescope is part of the Australia Telescope National Facility (\url{https://ror.org/05qajvd42}). Operation of ASKAP is funded by the Australian Government with support from the National Collaborative Research Infrastructure Strategy. ASKAP uses the resources of the Pawsey Supercomputing Research Centre. Establishment of ASKAP, Inyarrimanha Ilgari Bundara, the CSIRO Murchison Radio-astronomy Observatory and the Pawsey Supercomputing Research Centre are initiatives of the Australian Government, with support from the Government of Western Australia and the Science and Industry Endowment Fund. This paper includes archived data obtained through the CSIRO ASKAP Science Data Archive, CASDA (\url{http://data.csiro.au}). This publication makes use of data products from the Wide-field Infrared Survey Explorer, which is a joint project of the University of California, Los Angeles, and the Jet Propulsion Laboratory/California Institute of Technology, funded by the National Aeronautics and Space Administration. This publication also makes use of data products from NEOWISE, which is a project of the Jet Propulsion Laboratory/California Institute of Technology, funded by the Planetary Science Division of the National Aeronautics and Space Administration. This research has made use of the NASA/IPAC Infrared Science Archive, which is funded by the National Aeronautics and Space Administration and operated by the California Institute of Technology.

\bibliographystyle{plainnat}
\bibliography{references}

\begin{thebibliography}{15}
\providecommand{\natexlab}[1]{#1}
\providecommand{\url}[1]{\texttt{#1}}
\expandafter\ifx\csname urlstyle\endcsname\relax
  \providecommand{\doi}[1]{doi: #1}\else
  \providecommand{\doi}{doi: \begingroup \urlstyle{rm}\Url}\fi

\bibitem[Anand(2026)]{anand2026largal}
Aditya Anand.
\newblock {LARGAL}: Learning latent representation of radio galaxies for
  efficient compression and improved classification.
\newblock Young Scientist Journal, Vanderbilt University, May 2026 issue.
  \url{https://www.vanderbilt.edu/youngscientistjournal/article/largal-learning-latent-representation-of-radio-galaxies-for-efficient-compression-and-improved-classification/},
  2026.

\bibitem[Gupta et~al.(2023{\natexlab{a}})Gupta, Hayder, Norris, Huynh,
  Petersson, Wang, Andernach, Koribalski, et~al.]{gupta2023galcam}
Nikhel Gupta, Zeeshan Hayder, Ray~P. Norris, Minh Huynh, Lars Petersson,
  X.~Rosalind Wang, Heinz Andernach, B{\"a}rbel~S. Koribalski, et~al.
\newblock Deep learning for morphological identification of extended radio
  galaxies using weak labels.
\newblock \emph{Publications of the Astronomical Society of Australia},
  40:\penalty0 e044, 2023{\natexlab{a}}.
\newblock \doi{10.1017/pasa.2023.46}.

\bibitem[Gupta et~al.(2023{\natexlab{b}})Gupta, Norris, Huynh, Hayder, and
  Petersson]{rgnetdata}
Nikhel Gupta, Ray Norris, Minh Huynh, Zeeshan Hayder, and Lars Petersson.
\newblock {RadioGalaxyNET} dataset -- extended radio galaxies.
\newblock CSIRO Data Access Portal, collection 61068. doi:
  \href{https://doi.org/10.25919/btk3-vx79}{10.25919/btk3-vx79},
  2023{\natexlab{b}}.

\bibitem[Gupta et~al.(2024)Gupta, Hayder, Norris, Huynh, and
  Petersson]{gupta2024rgnet}
Nikhel Gupta, Zeeshan Hayder, Ray~P. Norris, Minh Huynh, and Lars Petersson.
\newblock {RadioGalaxyNET}: Dataset and novel computer vision algorithms for
  the detection of extended radio galaxies and infrared hosts.
\newblock \emph{Publications of the Astronomical Society of Australia},
  41:\penalty0 e001, 2024.
\newblock \doi{10.1017/pasa.2023.64}.

\bibitem[Hotan et~al.(2021)Hotan, Bunton, Chippendale, Whiting, Tuthill, Moss,
  et~al.]{hotan2021}
A.~W. Hotan, J.~D. Bunton, A.~P. Chippendale, M.~Whiting, J.~Tuthill, V.~A.
  Moss, et~al.
\newblock {Australian} {Square} {Kilometre} {Array} {Pathfinder}: {I}. system
  description.
\newblock \emph{Publications of the Astronomical Society of Australia},
  38:\penalty0 e009, 2021.
\newblock \doi{10.1017/pasa.2021.1}.

\bibitem[Jocher et~al.(2026)Jocher, Qiu, Liu, Lyu, Akyon, and
  Kalfaoglu]{ultralytics}
Glenn Jocher, Jing Qiu, Mengyu Liu, Shuai Lyu, Fatih~Cagatay Akyon, and
  Muhammet~Esat Kalfaoglu.
\newblock Ultralytics {YOLO26}: Unified real-time end-to-end vision models.
\newblock \emph{arXiv preprint arXiv:2606.03748}, 2026.
\newblock \doi{10.48550/arXiv.2606.03748}.

\bibitem[Lang(2014)]{lang2014}
Dustin Lang.
\newblock {unWISE}: Unblurred coadds of the {WISE} imaging.
\newblock \emph{The Astronomical Journal}, 147:\penalty0 108, 2014.
\newblock \doi{10.1088/0004-6256/147/5/108}.

\bibitem[Lin et~al.(2014)Lin, Maire, Belongie, Hays, Perona, Ramanan,
  Doll{\'a}r, and Zitnick]{lin2014coco}
Tsung-Yi Lin, Michael Maire, Serge Belongie, James Hays, Pietro Perona, Deva
  Ramanan, Piotr Doll{\'a}r, and C.~Lawrence Zitnick.
\newblock Microsoft {COCO}: Common objects in context.
\newblock In \emph{European Conference on Computer Vision (ECCV)}, volume 8693
  of \emph{Lecture Notes in Computer Science}, pages 740--755, 2014.
\newblock \doi{10.1007/978-3-319-10602-1_48}.

\bibitem[Marocco et~al.(2021)Marocco, Eisenhardt, Fowler, Kirkpatrick, Meisner,
  Schlafly, et~al.]{marocco2021}
Federico Marocco, Peter R.~M. Eisenhardt, John~W. Fowler, J.~Davy Kirkpatrick,
  Aaron~M. Meisner, Edward~F. Schlafly, et~al.
\newblock The {CatWISE2020} catalog.
\newblock \emph{The Astrophysical Journal Supplement Series}, 253:\penalty0 8,
  2021.
\newblock \doi{10.3847/1538-4365/abd805}.

\bibitem[Norris et~al.(2021)Norris, Marvil, Collier, Kapi{\'n}ska, O'Brien,
  Rudnick, et~al.]{norris2021}
Ray~P. Norris, Joshua Marvil, J.~D. Collier, Anna~D. Kapi{\'n}ska, Andrew~N.
  O'Brien, L.~Rudnick, et~al.
\newblock The {Evolutionary} {Map} of the {Universe} pilot survey.
\newblock \emph{Publications of the Astronomical Society of Australia},
  38:\penalty0 e046, 2021.
\newblock \doi{10.1017/pasa.2021.42}.

\bibitem[Peng et~al.(2025)Peng, Li, Wu, Zhang, Sun, and Wu]{peng2024dfine}
Yansong Peng, Hebei Li, Peixi Wu, Yueyi Zhang, Xiaoyan Sun, and Feng Wu.
\newblock {D-FINE}: Redefine regression task of {DETRs} as fine-grained
  distribution refinement.
\newblock In \emph{International Conference on Learning Representations
  (ICLR)}, 2025.

\bibitem[Robinson et~al.(2026)Robinson, Robicheaux, Popov, Ramanan, and
  Peri]{rfdetr}
Isaac Robinson, Peter Robicheaux, Matvei Popov, Deva Ramanan, and Neehar Peri.
\newblock {RF-DETR}: Neural architecture search for real-time detection
  transformers.
\newblock In \emph{International Conference on Learning Representations
  (ICLR)}, 2026.

\bibitem[Saakyan and Solntsev(2026)]{saakyan2026dfineseg}
Argo Saakyan and Dmitry Solntsev.
\newblock {D-FINE-seg}: Object detection and instance segmentation framework
  with multi-backend deployment.
\newblock \emph{arXiv preprint arXiv:2602.23043}, 2026.

\bibitem[Wright et~al.(2010)Wright, Eisenhardt, Mainzer, Ressler, Cutri,
  Jarrett, et~al.]{wright2010}
Edward~L. Wright, Peter R.~M. Eisenhardt, Amy~K. Mainzer, Michael~E. Ressler,
  Roc~M. Cutri, Thomas Jarrett, et~al.
\newblock The {Wide-field} {Infrared} {Survey} {Explorer} ({WISE}): Mission
  description and initial on-orbit performance.
\newblock \emph{The Astronomical Journal}, 140:\penalty0 1868--1881, 2010.
\newblock \doi{10.1088/0004-6256/140/6/1868}.

\bibitem[Zhang et~al.(2023)Zhang, Li, Liu, Zhang, Su, Zhu, Ni, and
  Shum]{zhang2023dino}
Hao Zhang, Feng Li, Shilong Liu, Lei Zhang, Hang Su, Jun Zhu, Lionel~M. Ni, and
  Heung-Yeung Shum.
\newblock {DINO}: {DETR} with improved denoising anchor boxes for end-to-end
  object detection.
\newblock In \emph{International Conference on Learning Representations
  (ICLR)}, 2023.

\end{thebibliography}

\end{document}